\documentclass[journal]{IEEEtran}
\ifCLASSINFOpdf
\else
\fi

\usepackage{times}
\usepackage{epsfig}
\usepackage{graphicx}
\usepackage{amsmath}
\usepackage{amssymb}
\usepackage{multirow}
\usepackage{notoccite}
\usepackage{multirow}
\usepackage{subfig}
\usepackage{tabularx}
\usepackage{color}

\usepackage{afterpage}

\newcommand{\black}[1]{\textcolor{black}{#1}}

\begin{document}
%
\title{Performance Characterization of Image Feature Detectors in Relation to the
Scene Content Utilizing a Large Image Database}
%
%
%

\author{Bruno Ferrarini$^1$\thanks{$^1$University of Essex, School of Computer Science and Electronic Engineering, Colchester CO4 3SQ, UK},
	Shoaib Ehsan$^1$,
	Ale\u{s} Leonardis$^2$\thanks{$^2$Dep. of Electrical Engineering, COMSATS Institute of Information Technology, Islamabad, Pakistan},
	Naveed Ur Rehman$^3$\thanks{$^3$School of Computer Science, University of Birmingham, Birmingham, UK},
	and Klaus D. McDonald-Maier$^1$
}

%
%

\markboth{IEEE TRANSACTIONS ON IMAGE PROCESSING,~Vol.~XX, No.~XX, XXXXX~20XX}%
{Ferrarini \MakeLowercase{\textit{et al.}}: Performance Characterization of Image Feature Detectors in Relation to the
Scene Content Utilizing a Large Image Database}
%



\maketitle

\begin{abstract}
Selecting the most suitable local invariant feature detector for a particular
application has rendered the task of evaluating feature detectors a critical issue in vision
research. Although the literature offers a variety of comparison works
focusing on performance evaluation of image feature detectors under several types of image
transformations, the influence of the scene content on the performance of local feature
detectors has received little attention so far. This paper aims to bridge this gap with
a new framework for determining the type of scenes which maximize and minimize the
performance of detectors in terms of repeatability rate. The results are presented for several
state-of-the-art feature detectors that have been obtained using a large image database of
20482 images under JPEG compression, uniform light and blur changes with 539 different
scenes captured from real-world scenarios. These results
provide new insights into the behavior of feature detectors.
\end{abstract}


%
\IEEEpeerreviewmaketitle

\section{Introduction}
\label{sec:intro}

Local feature detection has been a challenging area of interest for the computer vision community for some time. A large
number of different approaches have been proposed so far,
thus making evaluation of image feature detectors an active
research topic in the last decade or so. Most 
evaluations available in the literature focus mainly on
characterizing feature detectors' performance under
different image transformations without analyzing 
the effects of the scene content in detail. In \cite{tissainayagam_assessing_2004}, the feature tracking
capabilities of some corner detectors are assessed utilizing
static image sequences of a few different scenes. Although
the results permit to infer a dependency of the detectors'
performance on the scene content, the methodology
followed is not specifically intended to highlight and formalize such a
relationship, as no classification is assigned to the scenes.
The comparison work in \cite{mikolajczyk_comparison_2005} gives a formal definition for
textured and structured scenes and shows the repeatability
rates of six feature detectors. The results provided by \cite{mikolajczyk_comparison_2005}
show that the content of the scenes influences the
repeatability but the framework utilized and the small
number of scenes included in the datasets \cite{mikolajczyk_oxford_dataset} do
not provide a comprehensive insight into the behavior of the
feature detectors with different types of scenes. In \cite{fraundorfer2005novel}, the
scenes are classified by the complexity of their 3D
structures in complex and planar categories. The
repeatability results reveal how detectors perform for those
two categories. The limit in the generality of the analysis
done in \cite{fraundorfer2005novel} is due to the small number and variety
of the scenes employed, whose content are mostly human-made.
This paper aims to help better understand the effect of the
scene content on the performance of several state-of-the art
local feature detectors. The main goal of this work is to
identify the biases of these detectors towards particular
types of scenes, and how those biases are affected by three
different types and amounts of transformations (JPEG
compression, blur and uniform light changes). The
methodology proposed utilizes the improved repeatability
criterion presented in \cite{ehsan_improved_2010}, as a measure of the performance
of feature detectors, and the large database \cite{ehsan_dataset} of images
consisting of 539 different real-world scenes containing a
wide variety of different elements. 
\black{This paper offers a more complete understanding of the evaluation framework first described in the conference version \cite{ferrarini2015performance}, providing further background, description, insight, analysis and evaluation.}\\
The remainder of the paper is organized as follows. Section \ref{sec:related_work}
provides an overview of the related work in the field of
feature detector evaluation and scene taxonomy. In Section
\ref{sec:evaluation_framework}, the proposed evaluation framework is described in detail.
Section \ref{sec:image_dataset} is dedicated to the description of the image
database utilized for the experiments. The results utilizing
the proposed framework are presented and discussed in
Section \ref{sec:results}. Finally, Section \ref{sec:conclusions} presents the conclusions.

\begin{figure*}[h!]
\centering
\includegraphics[width=14cm,keepaspectratio=true]{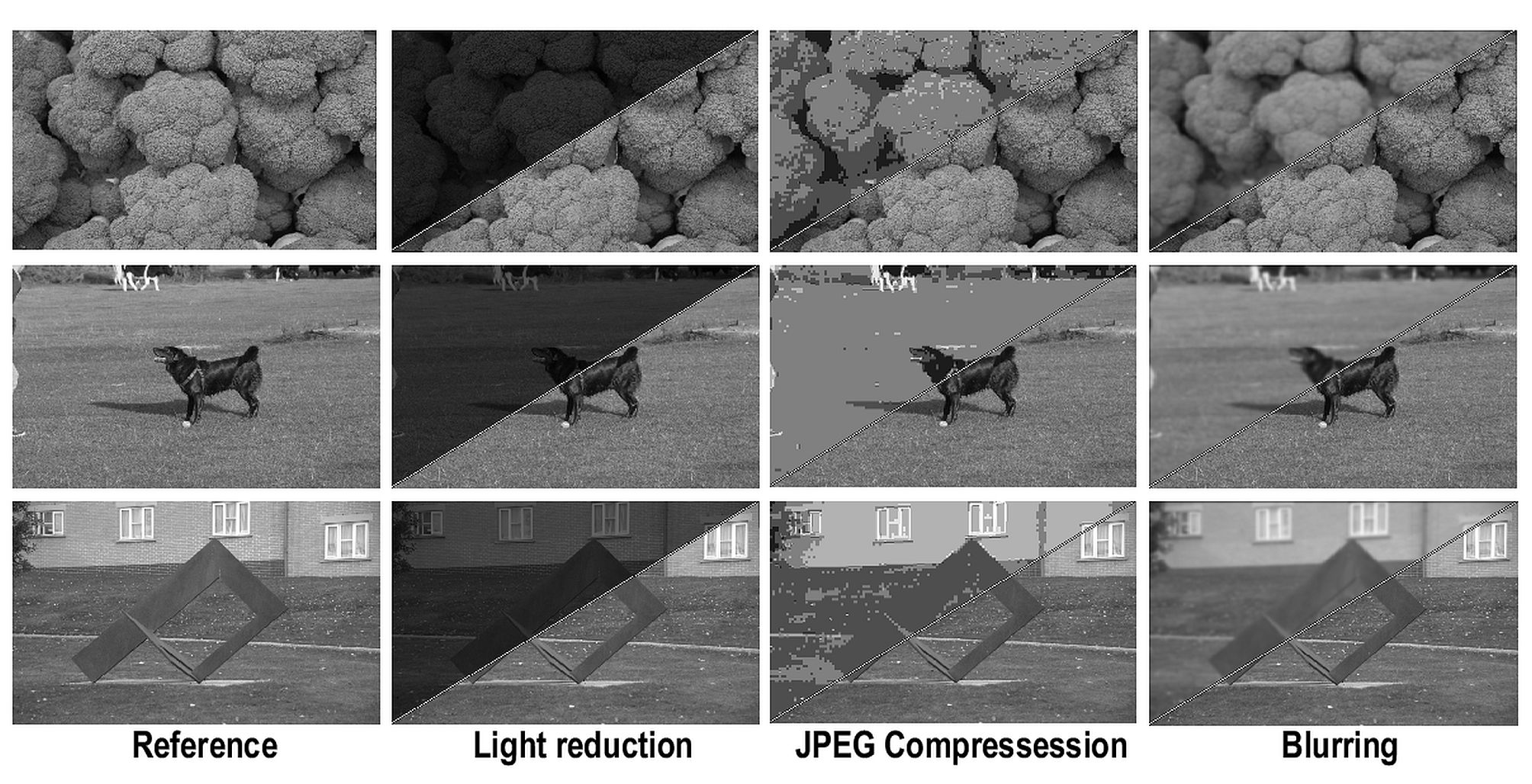}
\caption{The reference image of three scenes and the effect of the application of $60\%$ of light reduction, $98\%$ of JPEG compression rate and $4.5\sigma$ Gaussian blur.}
\label{fig:transformation_sample}
\end{figure*}

\section{Related Work}
\label{sec:related_work}

The contributions to the evaluation of local feature
detectors are numerous and vary based on: 1) the metric used
for quantifying the detector performance, 2) the
framework/methodology followed and 3) the image databases
employed. Repeatability is a desirable property for feature
detectors as it measures the grade of independence of the
feature detector from changes in the imaging conditions.
For this reason, it is frequently used as a measure of
performance of local feature detectors. A definition of
repeatability is given in \cite{schmid_evaluation_2000} where, together with the
information content, it is utilized as a metric for comparing
six feature detectors. A refinement of the definition of
repeatability is given in \cite{mikolajczyk_scale_2004}, and used for assessing six state-of-the-art feature detectors in \cite{mikolajczyk_comparison_2005} under several types of
transformations on textured and structured scenes. 
\black{Two criteria for improved repeatability measure are introduced in \cite{ehsan_improved_2010} that provide results which are more consistent with the actual performance of several popular feature detectors on the
widely-used Oxford datasets \cite{mikolajczyk_oxford_dataset}. The improved repetability criteria are employed in the evaluation  framework proposed in \cite{ferrarini2016performance} and in \cite{ehsan2016generic}, which presents a method to assess the performance bounds of detectors.}
Moreover,
repeatability is used as a metric for performance evaluation
in \cite{fraundorfer_evaluation_2004} and \cite{fraundorfer2005novel} that utilize non-planar, complex and simple
scenes.

The performance of feature detectors has also been
assessed employing metrics other than repeatability. The
performance measure in \cite{dickscheid_coding_2011} is completeness, while feature
coverage is used as a metric in \cite{ehsan_rapid_2013}. The feature detectors
have also been evaluated in the context of specific
applications, such as in \cite{tissainayagam_assessing_2004}, where corner detectors are
assessed in the context of point feature tracking
applications.

\begin{figure*}[t!]
\centering
\includegraphics[width=15cm,keepaspectratio=true]{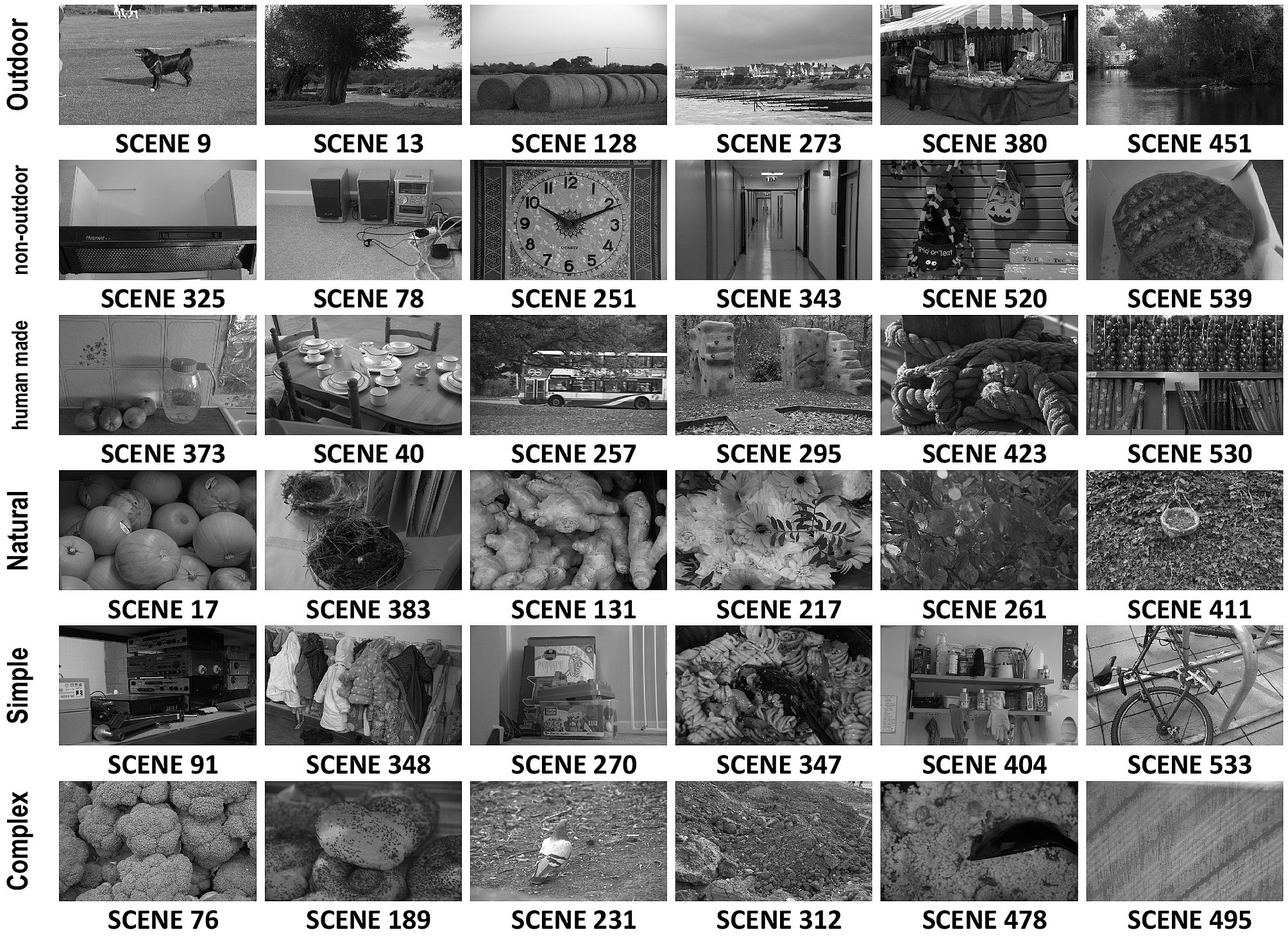}
\caption{Some images from the database utilized for the experiments. Each row shows images belonging to the same category: outdoor, non-outdoor, human-made content, natural content, simple and complex edges.}
\label{fig:sample}
\end{figure*}

\section{The Proposed Evaluation Framework}
\label{sec:evaluation_framework}





The proposed framework has been designed by keeping
in mind the objective of evaluating the influence of scene
content on the performance of a wide variety of state-of-the-art feature detectors. A proper application of such a
framework requires a large image database $I$ organized in a
series of $n$ datasets. Each dataset needs to contain images
from a single scene with different amounts of image
transformation. The images included in such a database
should be taken from a large variety of different real-world
scenarios. The proposed framework consists of the steps
discussed below.

\subsection{Repeatability data}
\label{sebsec:repeatability_data}

The framework is based on the repeatability criterion
described in \cite{ehsan_improved_2010}, whose consistency with the actual
performance of a wide variety of feature detectors has been
proven across well-established datasets \cite{mikolajczyk_oxford_dataset}. As proposed in
\cite{ehsan_improved_2010}, the repeatability rate is defined as follows:

 \begin{equation}
 \label{eq:rep}
 Repeatability = {\frac{N_{rep}}{N_{ref}}}
 \end{equation}
 
where $N_{rep}$ is the total number of repeated features and
$N_{ref}$ is the number of interest points in the common part of
the reference image.
Let $A$ and $P$ be the sets of indices representing the $m$
discrete amount of transformation and the scenes
respectively.

 \begin{equation}
 \label{eq:A}
 A = \left \{1,2,3,... ... m\right \}
 \end{equation}

  \begin{equation}
 \label{eq:P}
 P = \left \{1,2,3,... ... n\right \}
 \end{equation}
where $m$ corresponds to the maximum amount of
transformation and $‘1’$ relates to the reference image (zero
transformation); $n$ is the total number of scenes and each
scene is utilized to build one separate dataset, thus finally
resulting in $n$ datasets in total. Let $B_{kd}$ be the set of
repeatability rates computed for step $k$ (corresponding to $k$
image transformation amount) for a feature detector $d$
across $n$ datasets (which implies repeatability values for $n$
scenes):

 \begin{equation}
 \label{eq:Bkd}
 B_{kd} = \left \{B_{1kd},B_{2kd}, ... ... B_{nkd}\right \}
 \end{equation}
 
Each set $B_{kd}$ contains $n$ repeatability ratios, one for each
dataset. In particular, for the image database utilized 
in this work for the experiments \cite{ehsan_dataset}, $n$ is 539
while maximum value of $k$ is 10 or 14 depending on which
transformation is considered. Thus, $B_{kd}$  includes 539
values of repeatability for the $k^{th}$ step.

\subsection{Scene rankings}
\label{fig:scene_rankings}

The top and 
lowest 
rankings for each detector $d$ are built
selecting the $j$ highest and the lowest repeatability scores at
$k$ amount of image transformation. Let $T_{kd}(j)$ and $W_{kd}(j)$
the sets containing the indices of the scenes whose
repeatability falls in the top and lowest ranking respectively:

 \begin{equation}
 \label{eq:Tkd}
 T_{kd}(j) = \left \{S_{kd(1)},S_{kd(1)}, ... ... S_{kd(j)}\right \}
 \end{equation}
 
  \begin{equation}
 \label{eq:Wkd}
 W_{kd}(j) = \left \{S_{kd(n)},S_{kd(n-1)}, ... ... S_{kd(n-j+1)}\right \}
 \end{equation}

where $S_{kd(i)} \in P$ is the scene index corresponding to the $i^{th}$
highest repeatability score obtained by the detector $d$ for the
scene under $k$ amount of transformation. Thus, in accordance
with this notation, $S_{kd(1)}$ is the scene for which the detector
scored the best repeatability score, $S_{kd(2)}$ corresponds to the
second highest repeatability rate, $S_{kd(3)}$ to the third highest
and so on, until $S_{kd(n)}$ which is for the lowest one.

\subsection{Scene classification}
\label{sec:scene_classification}

The scenes are attributed with three labels on the basis of
human judgment. As described in Table \ref{table:class}, each label is
dedicated to a particular property of the scene and has been
assigned independently from the others. These attributes
are: the location type $(f)$, which may take the label outdoor
or indoor, the type of the elements contained $(g)$, which may
take the label natural or human-made, and the perceived
complexity of the scene $(h)$, which may take the label simple
or complex. Figure \ref{fig:sample} shows a sample of the scenes from the
image database \cite{ehsan_dataset} utilized for the experiments grouped so that each row shows scenes sharing the same value for one of the three labels $f$, $g$ and $h$. Scene 9
is tagged as outdoor and, along with scene 76 and 17,
contains natural elements. The scenes 40, 530 and 373 are
labeled as human-made and the first is also classified
as indoor. The scene 530 is categorized as a simple scene as
it includes a few edges delimiting well contrasted areas.
Although the main structures (broccolis borders) can be
identified in scene 76, the rough surface of the broccolis is
information rich that results in labeling this scene as
complex.


\begin{table}[h!]
\centering
\begin{tabular}{|l|p{1.7cm}|p{4cm}|}

\hline

\multirow{2}{*}{\begin{tabular}[c]{@{}c@{}}Location \\ Type\end{tabular}} & Outdoor & Indoor scene and close-up a single or of a few objects. \\ \cline{2-3} 

 & Indoor & The complement of above. \\ \hline

\multirow{2}{*}{\begin{tabular}[c]{@{}c@{}}Object \\ Type\end{tabular}} & Human-made & Elements are mostly artificial. \\ \cline{2-3} 

 & Natural & Elements are mostly natural. \\ \hline

\multirow{2}{*}{Complexity} & Simple & A few edges with quite regular shapes. \\ \cline{2-3} 

 & Complex & A large number of edges with fractal-like shapes. \\ \hline

\end{tabular}

\caption[Classification labels and criteria]{Classification labels and criteria}
\label{table:class}
\end{table}


\subsection{Ranking trait indices}
\label{sec:ranking_traits_indices}

The labels of the scenes included in the rankings, (\ref{eq:Tkd}) and
(\ref{eq:Wkd}), are examined in order to determine the dominant types
of scenes. For each ranking $T_{kd}(j)$ and $W_{kd}(j)$, the ratios of
scenes classified as \textit{outdoor}, \textit{human-made} and \textit{simple} are
computed. Thus, three ratios are associated to each ranking
where higher values mean higher share of the scene type
associated:
 \begin{equation}
 \label{eq:T_I}
 \forall S_{i}\in T_{kd}: T_{kd}.[F,G,H]=\frac{\sum S_{i}.[f,g,h]}{j} 
 \end{equation}
 
  \begin{equation}
 \label{eq:W_I}
 \forall S_{i}\in W_{kd}: W_{kd}.[F,G,H]=\frac{\sum S_{i}.[f,g,h]}{j} 
 \end{equation}

 These vectors contain three measures which represent the
extent of the bias of detectors. For example, if a top ranking
presents $F = 0.1$, $G = 0.25$ and $H = 0.8$, it can be
concluded that the detector, for the given amount of image
transformation, works better with scenes where its' element are
mostly natural (low $G$), with simple edges (high $H$) and that
are not outdoor (low $F$). As opposed to that, if the same
indices were for a lowest ranking it could be concluded that
the detector obtains its lowest results for non-outdoor ($F$)
and natural ($G$) scenes with low edge complexity ($H$).


\section{Image Dataset}
\label{sec:image_dataset}

The image database used for the experiments is discussed
in this section and is available at \cite{ehsan_dataset}. It contains a large
number of images, 20482, from real-world scenes. This database
includes a wide variety of outdoor and indoor scenes
showing both natural and human-made elements. The
images are organized in three groups of datasets,
corresponding to the three transformations: JPEG
compression, uniform light and Gaussian blur changes.
Each dataset includes a reference image for the particular
scene and several images of the same scene with different
amounts of transformation for a total of $10$ images for
Gaussian blur and $14$ for JPEG compression and uniform
light change transformations. Figure \ref{fig:sample} provides both a
sample of the scenes available and an example of
transformation applied to an image.

Several well-established datasets, such as \cite{mikolajczyk_oxford_dataset}, are
available for evaluating local feature detectors, however are not
suitable for use with the proposed framework due to the
relatively small number and lesser variety of scenes included,
and the limited range of the amount of transformations
applied. For example, UBC dataset \cite{mikolajczyk_oxford_dataset}, which was used in
\cite{mikolajczyk_comparison_2005}, includes images for JPEG compression ratios varying
from $60\%$ to $98\%$ only. Among the Oxford datasets \cite{mikolajczyk_oxford_dataset},
Leuven offers images under uniform light changes, however the
number of images in that dataset is only six. Although the
database employed in \cite{aana_es_interesting_2012} for assessing several feature
detectors under different light conditions contains a large
number of images, the number of scenes are limited to $60$.
Moreover, these scenes were captured in a highly controlled
environment so they are lesser representative of real-world scenario in comparison of the image database used here.

The images included in the database utilized for this
work have a resolution of $717\times 1080$ pixels and consist of
$539$ real-world scenes. Each transformation is applied in
several discrete steps to each of the scenes. The Gaussian
blur amount is varied in 10 discrete steps from $0$ to $4.5\sigma$ ($10\times 539 = 5390$ images), JPEG compression ratio is increased
from $0$ to $98\%$ in $14$ steps. Similarly, the amount of light is
reduced from $100\%$ to $10\%$ ($14\times 539 = 7546$ images).
Thus, the database includes a dataset of 10 or 14 images for
each of the $539$ scenes for a total of $20482$ images. The
ground truth homography that relates any two images of the
same scene is a $3\times 3$ identity matrix for all three
transformations as there is no geometric change involved.

Accordingly with the classification criteria introduced in the Section \ref{sec:scene_classification},  $51\%$ of the $539$ scenes have been labeled as outdoor, $65\%$ contain mostly human made elements and $51\%$ have been attributed with the simple label. Overall, the database has reasonable balance between the content types introduced by the proposed classification criteria, so that it becomes possible to produce significant bias descriptors for the local feature detectors. 
\section{Results}
\label{sec:results}
The proposed framework has been applied for producing
the top and lowest rankings for a set of eleven feature
detectors which are representative of a wide variety of
different approaches \cite{tuytelaars_local_2008} and includes the following: Edge-Based Region (EBR) \cite{tuytelaars_content-based_1999}, Harris-Affine (HARAFF),
Hessian-Affine (HESAFF) \cite{mikolajczyk_affine_2002}, Maximally Stable
Extremal Region (MSER) \cite{matas_robust_2004}, Harris-Laplace (HARLAP),
Hessian-Laplace (HESLAP) \cite{mikolajczyk_scale_2004}, Intensity-Based Regions
(IBR) \cite{tuytelaars_matching_2004}, SALIENT \cite{kadir_affine_2004}, Scale-invariant Feature
Operator (SFOP) \cite{forstner_detecting_2009}, Speeded-Up Robust Feature (SURF)
\cite{baya_speeded-up_2008} and SIFT \cite{lowe_object_1999}.\\
The first subsection provides details
on how the repeatability data have been obtained and the second one is dedicated to the discussion
about the ranking traits of each local invariant feature detector.
\subsection{Repeatability Data}
\label{sec:repeatability_data}
The repeatability data are obtained for each transformation
type utilizing the image database discussed in Section \ref{sec:image_dataset}.
This data is collected using the authors' original programs
with control parameter values suggested by the respective authors. The
feature detector parameters could be varied in order to
obtain a similar number of extracted features for each
detector. However, this has a negative impact on the
repeatability of a detector \cite{mikolajczyk_scale_2004} and is therefore not desirable
for such an evaluation. \\
Utilizing the dataset described in detail in Section \ref{sec:image_dataset}, 
$18865$ repeatability rates have been
computed for each local feature detector with the exception of SIFT, 
which has been assessed only under JPEG compression.
It should be noted that SIFT detects more than 20,000 features for some images in the image database which makes it very time-consuming to do such a detailed analysis for SIFT. In the case of JPEG image database, it took more than two months to obtain results on HP ProLiant DL380 G7 system with Intel Xeon 5600 series processors. Therefore, results for SIFT are not provided in this section.\\
The number of datasets is $539$, the number of discrete step of transformation 
amount $k$ varies across the transformations considered. We employed: $k=14$ for JPEG compression
and uniform light change transformations and $k=10$ for Gaussian blur. Since the 
first step of transformation amount corresponds
to the reference image, the number of set $B_{kd}$ (\ref{eq:Bkd}) is 13 for 
JPEG compression and light changes, and 9 for blurring for a total
of $2\times(13 \times 539) + 9 \times 539 = 18865$ repeatability 
rate values for each detector.

\subsection{Trait Indices}
\label{subsec:traits}

In this section the trait indices for all the assessed image feature detectors are presented and discussed. The trait indices have been designed to provide a measure of the bias of the feature detector for any of the types of scene introduced by the classification criterion described in the Section \ref{sec:scene_classification}. In other words, they are indicative of the types of scene for which a feature detector is expected to perform well.  
Accordingly, with the definition provided in the Section \ref{sec:ranking_traits_indices}, they represent the percentage of the scenes in the top and lowest rankings of a particular type of scene. Thus, they permit to characterize quantitatively the performance of feature detectors from the point of view of the scene content.\\
The trait indices are built starting from the top and lowest rankings of any feature detector. For obtaining the results presented in this work, the evaluation framework has been applied utilizing a ranking length of 20 ($j=20$). Finally, the related trait indices are computed by applying the equations (\ref{eq:T_I}) and (\ref{eq:W_I}) presented in the Section \ref{sec:ranking_traits_indices}. \\
The results of all detectors are shown in the Figures \ref{fig:ebr_tables}--\ref{fig:surf_tables} and discussed below. The results are presented utilizing radar charts: the transformation amounts are shown on the external perimeter and increase clockwise; the trait indices are expressed in percentage with the value which increases from the center (\%0) to the external perimeter (\%100) of the chart. \\

\subsubsection{EBR trait indices}
\label{subsec:ebr_traits}

\begin{figure}[!tb]
\centering
\includegraphics[width=8.5cm,keepaspectratio=true]{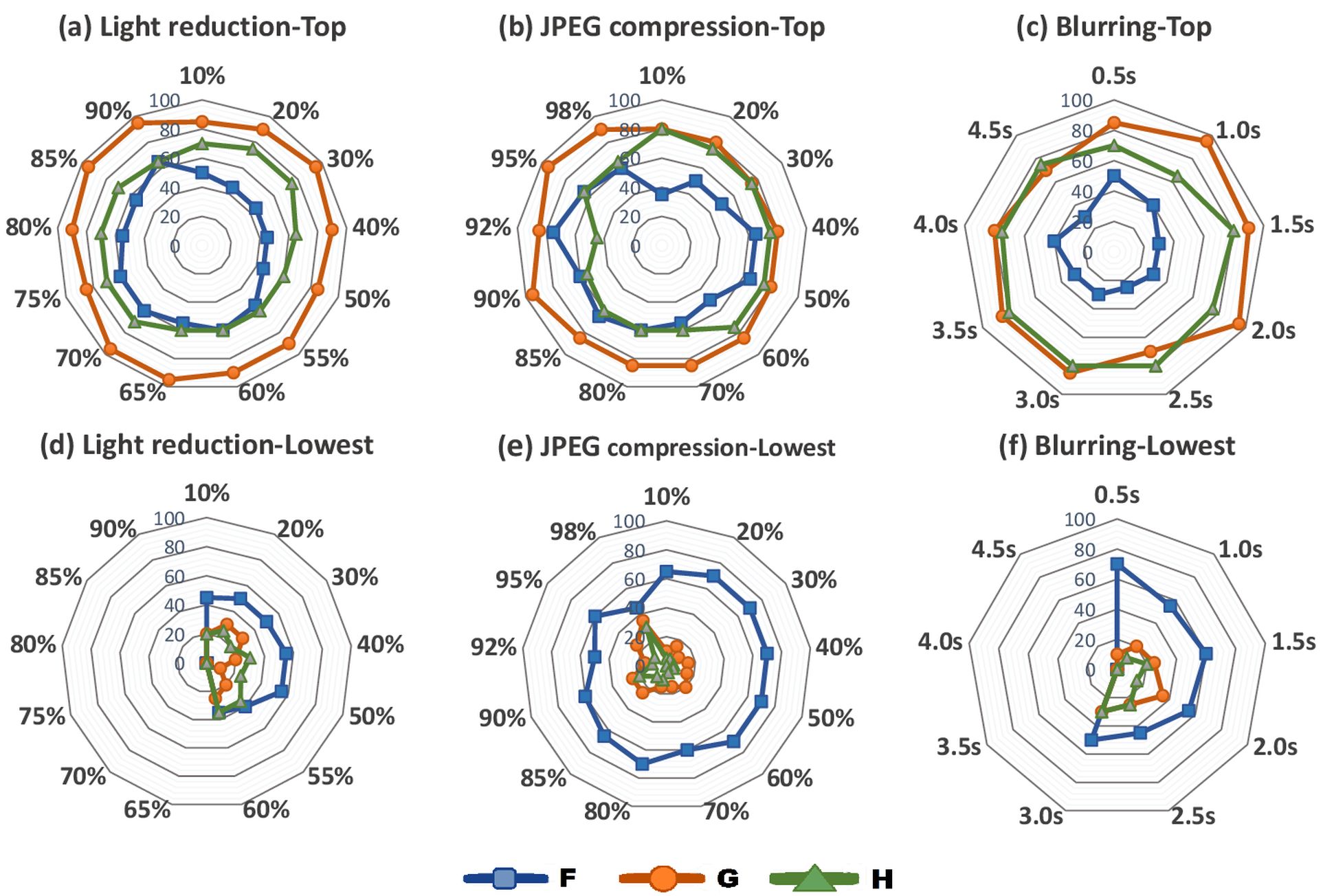}
\caption{
Top and lowest trait indices of EBR in percentage for different amounts of light reduction (a,d), JPEG compression (b,e) and blurring (c,f).
}
\label{fig:ebr_tables}
\end{figure}

All the available trait indices of Edge-Based Regions (EBR) detector \cite{tuytelaars_content-based_1999} are reported in Figure \ref{fig:ebr_tables}. 
\black{The performances of EBR are very sensitive to uniform light reduction and Gaussian blur \cite{ferrarini2016performance}. In particular, for light reduction higher than $60\%$  and Guassian blur equal or greater than  $3.0\sigma$, there are more than 20 scenes for which EBR scores a repeatability rate value of $0\%$ making impossible to form a scene ranking as described above (Section \ref{subsec:ebr_traits}). For this reason, in Figures \ref{fig:ebr_tables}.d and  \ref{fig:ebr_tables}.e the trait indices for the higher amounts of those transformation are omitted.}\\
EBR exhibits high values (around $80\%$ - $90\%$) of $G$ in the top rankings and low (rarely above $25\%$) in the lowest rankings, denoting a strong bias towards the scenes including many human made elements. EBR performs generally well on simple scenes as well, in particular under Gaussian blur, with the share of simple scenes never below $70\%$. The values assumed by $F$ indices are not indicative of the EBR's bias for particular location types as they assume very similar values between the top and lowest rankings.\\

\subsubsection{HARLAP and HARAFF trait indices}
\begin{figure}[!tb]
\centering
\includegraphics[width=8.5cm,keepaspectratio=true]{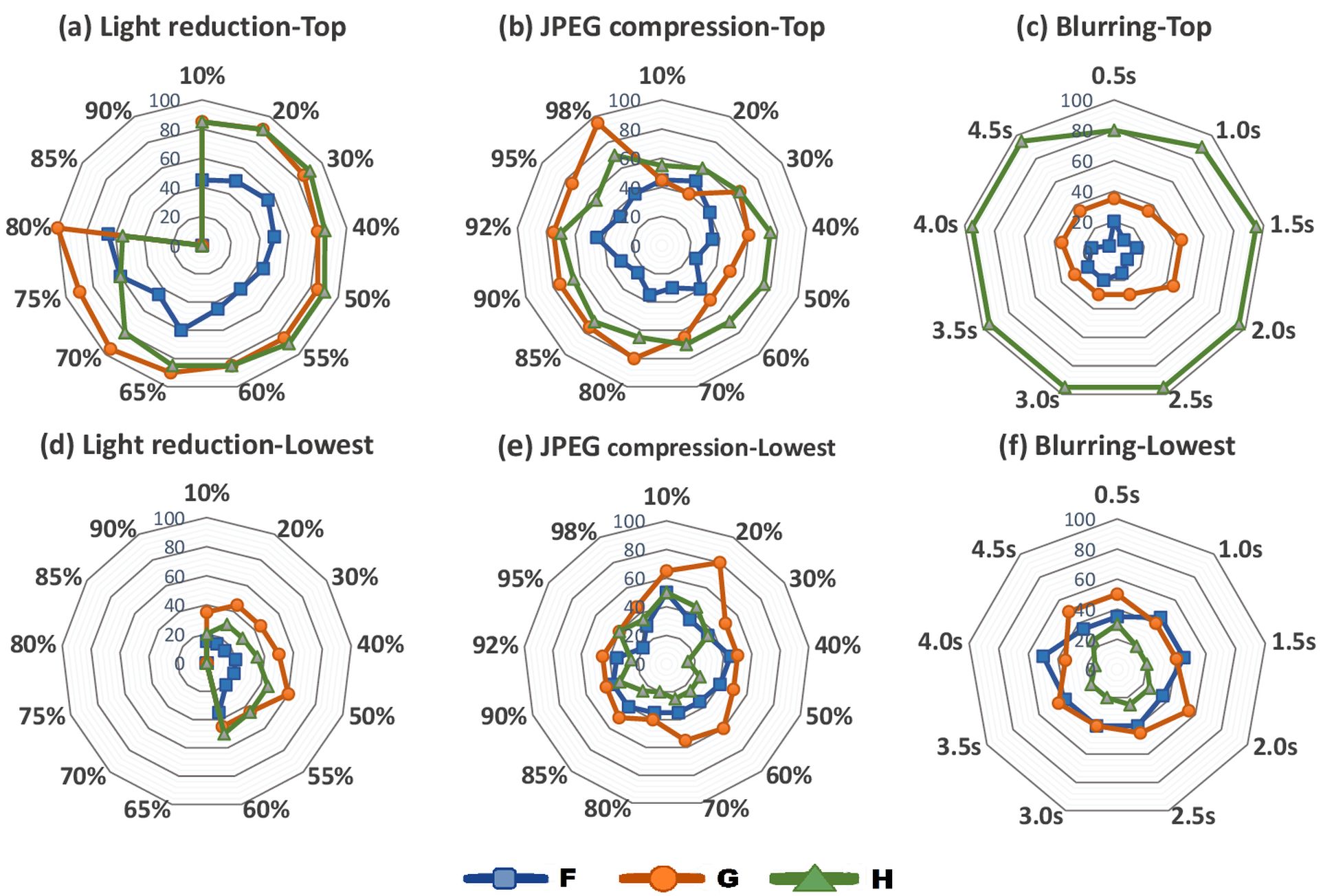}
\caption{
Top and lowest trait indices of HARLAP in percentage for different amounts of light reduction (a,d), JPEG compression (b,e) and blurring (c,f).
}
\label{fig:harlap_tables}
\end{figure}

\begin{figure}[!tb]
\centering
\includegraphics[width=8.5cm,keepaspectratio=true]{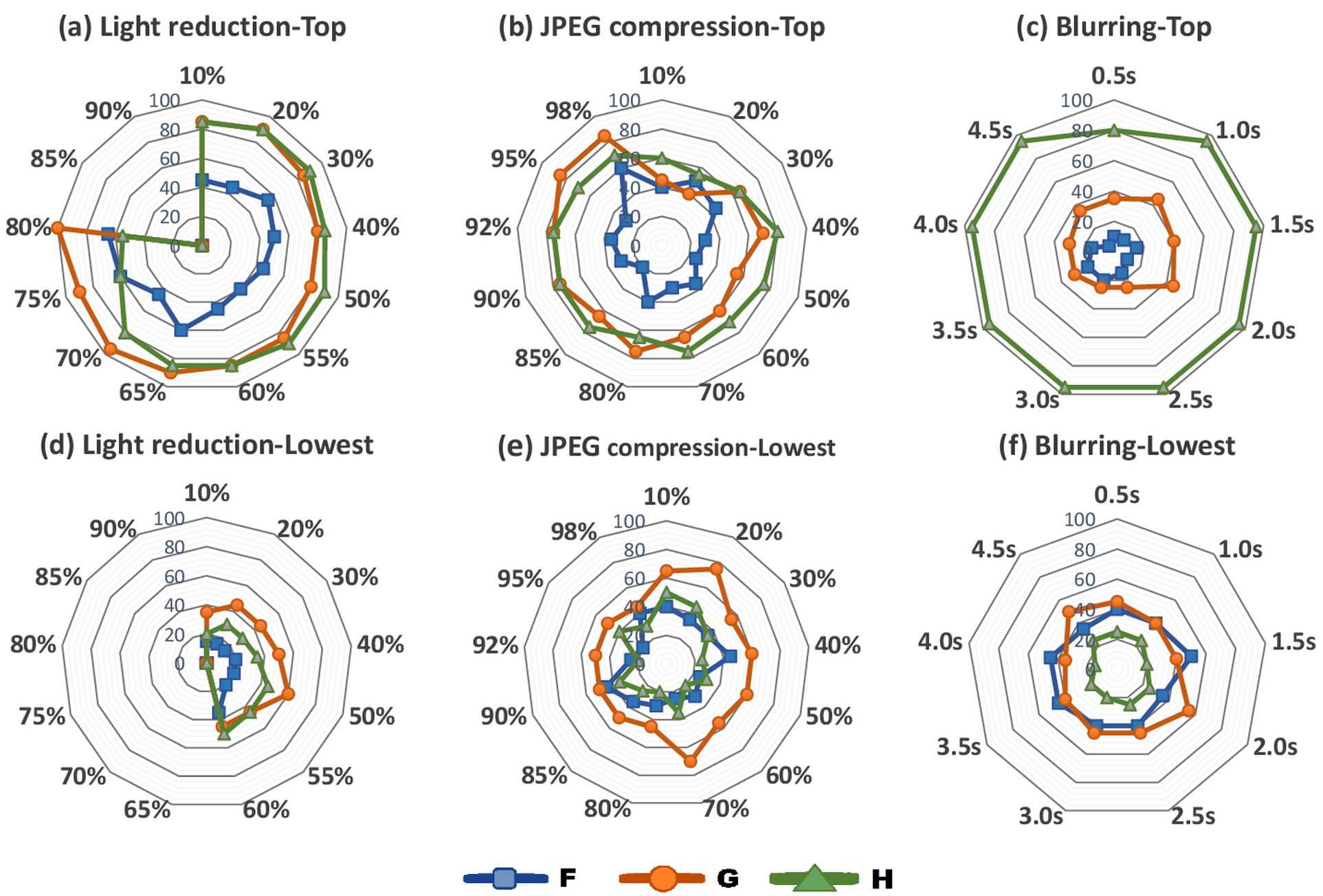}
\caption{
Top and lowest trait indices of HARAFF in percentage for different amounts of light reduction (a,d), JPEG compression (b,e) and blurring (c,f).
}
\label{fig:haraff_tables}
\end{figure}

The rankings of HARLAP \cite{mikolajczyk_scale_2004} and HARAFF
\cite{mikolajczyk_affine_2002} are very similar to each other and so are the values of their trait indices. Both of them are particularly prone to uniform light changes and the trait indices for high level of transformation amounts are not available. Figures \ref{fig:harlap_tables}.a and \ref{fig:haraff_tables}.a  report the results for the top rankings up to $80\%$ of light reduction and up to $60\%$ for the lowest rankings.\\
HARLAP and HARAFF present a bias towards simple scenes, which is particularly strong under uniform light reduction and blurring as can be inferred by the high values, which  $H$ assumes in the top twenties and the low values in the relative lowest rankings. A clear preference of those detectors for human made objects can be claimed under light changes, however this is not the case under JPEG compression and Gaussian blur whose related  $G$ indices are too close between top and lowest rankings to draw any conclusion. The $F$ indices are extremely low (never above $20\%$) for the top twenty rankings under Gaussian blur revealing that HARLAP and HARAFF deal better with non-outdoor scenes under this particular transformation.\\


\subsubsection{HESLAP and HESAFF trait indices}

\begin{figure}[!tb]
\centering
\includegraphics[width=8.5cm,keepaspectratio=true]{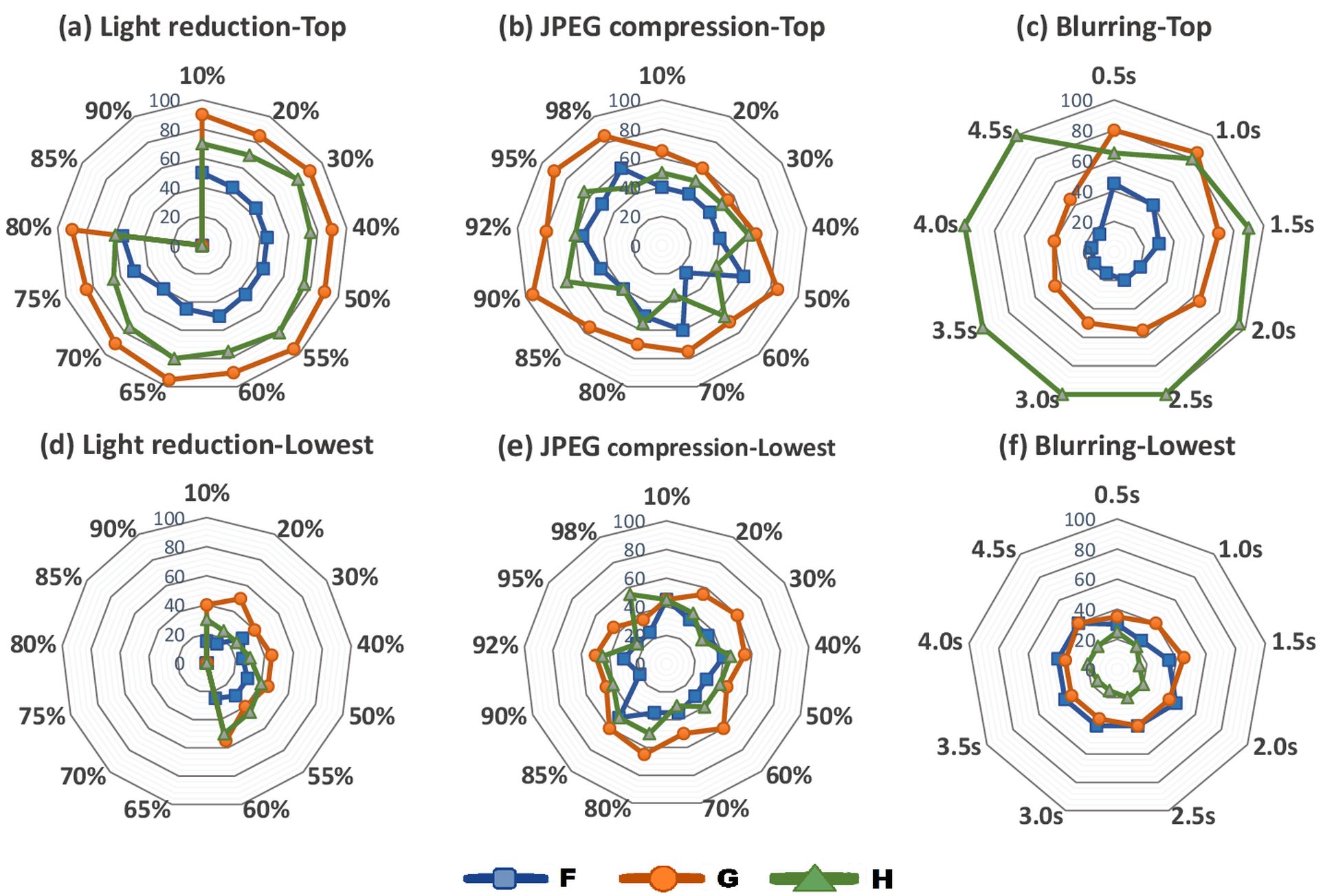}
\caption{
Top and lowest trait indices of HESLAP in percentage for different amounts of light reduction (a,d), JPEG compression (b,e) and blurring (c,f).
}
\label{fig:heslap_tables}
\end{figure}

\begin{figure}[!tb]
\centering
\includegraphics[width=8.5cm,keepaspectratio=true]{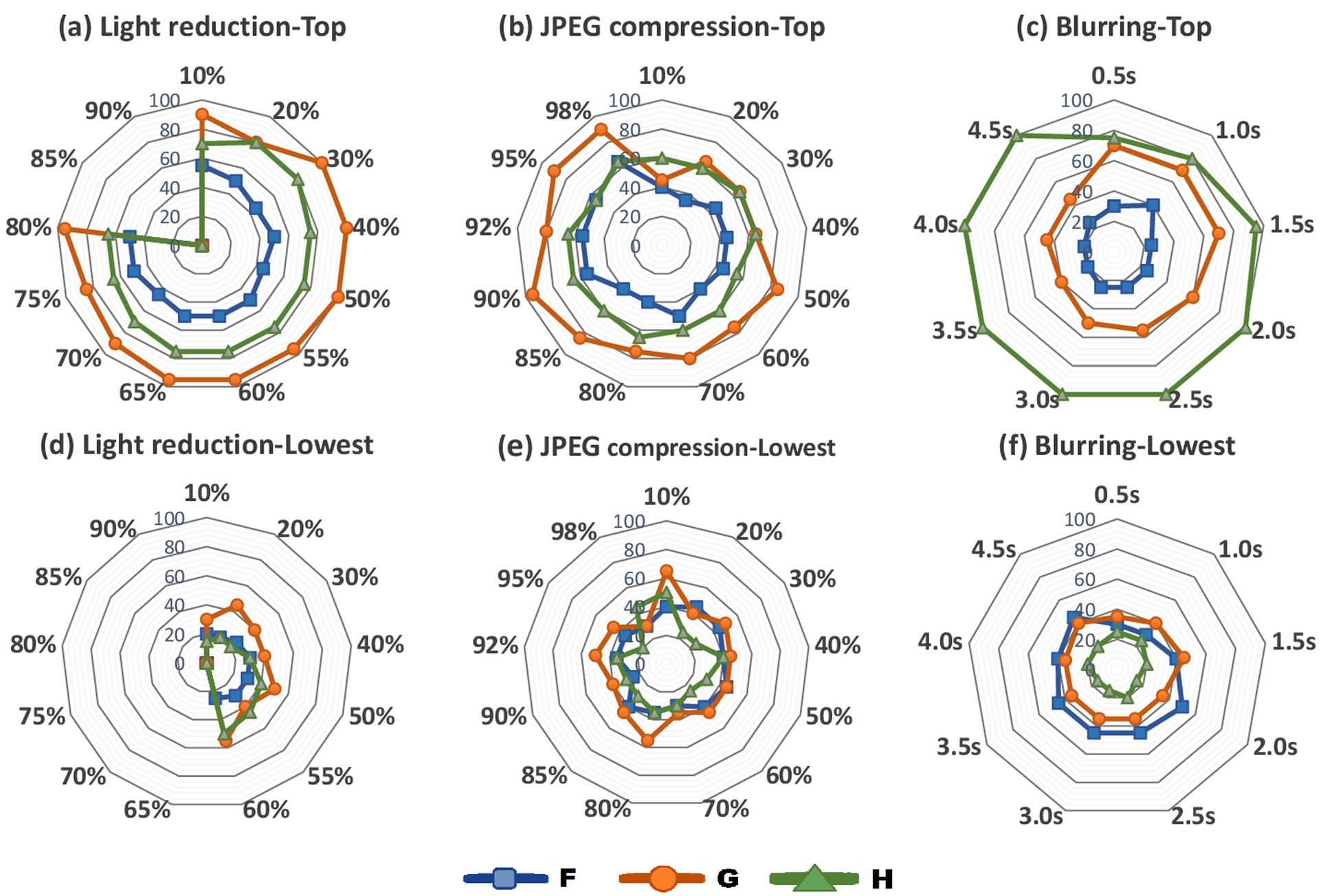}
\caption{
Top and lowest trait indices of HESAFF in percentage for different amounts of light reduction (a,d), JPEG compression (b,e) and blurring (c,f).
}
\label{fig:hesaff_tables}
\end{figure}

Due to the similarities between the approach in localizing the interest point in images, HESLAP
and HESAFF present many similarities between their trait indices. Similarly to HARLAP and HARAFF, uniform light changes have a strong impact on the HESLAP and HESAFF's performance \cite{ferrarini2016performance}. For that reason, the Figures \ref{fig:heslap_tables}.a and \ref{fig:hesaff_tables}.a show only the results for the top rankings of up to $80\%$ of light reduction and up to $60\%$ for the lowest rankings.\\
HESLAP and HESAFF perform better on scenes characterized by simple elements and edges under blurring (especially for high $\sigma$ values) and uniform light decreasing. The same indices, $H$, computed under JPEG compression present fluctuations around $50\%$ for both the top and lowest rankings without \black{ bending towards simple or complex scenes}. Both the detectors perform well on scenes containing human-made  elements under light reduction, JPEG compression and up to $2.5\sigma$ of Gaussian blur. Although both HESLAP and HESAFF do not have any bias for outdoor scenes, the HASLAP's $F$ index decrease from $45\%$ to $15\%$  constantly over the variation range of blurring amount. \\

\subsubsection{SIFT}

\begin{figure}[!b]
\centering
\includegraphics[width=8.5cm,keepaspectratio=true]{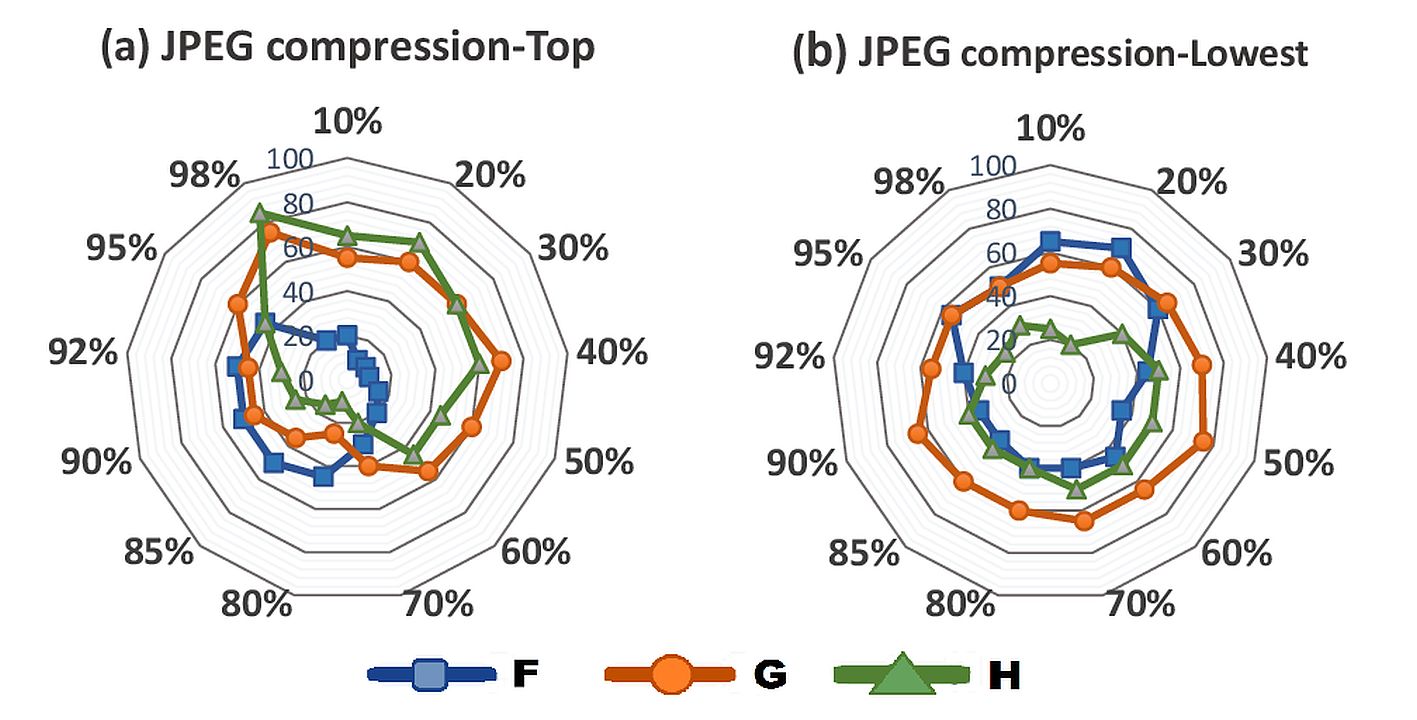}
\caption{
Top and lowest trait indices of SIFT in percentage for different amounts JPEG compression (a,b).
}
\label{fig:sift_tables}
\end{figure}

From the trait index data, it is not possible to determine a clear bias in the performance of SIFT \cite{lowe_object_1999}, as the values of the trait indices fluctuate over the entire range of the JPEG compression rate.
\black{
Figure \ref{fig:sift_tables} confirms a bias towards simple and human made objects only between $10\%$ and $60\%$ and at at $98\%$ of JPEG compression. whereas the indices $G$ and $H$ present large fluctuations in the top twenty scene rankings for the other compression rates. In particular, between $70\%$ and $90\%$ of JPEG compression their values are significantly lower than ones at other compression amounts and reach a minimum at $80\%$ which are $10\%$ for $H$ and $25\%$ for $G$.
Similar variations can be appreciated also for $F$ in both top and lowest rankings with values variations broad up to $40\%$. While the $G$ and $H$ indices in many cases present small differences between the top and lowest rankings, the $F$ indices are often inversely related. For example, at $30\%$ compression, $F$ is equal to $10\%$ for the top twenty and $60\%$ for the bottom twenty, $G$ is $60\%$ in both cases and $H$ differs for just $20\%$ between the top and lowest rankings.\\
}
In conclusion, the classification criteria adopted in this work permits to infer a strong dependency of SIFT on the JPEG compression rate variations, however, it does not allow to draw any conclusions about the general bias, if any exists,  towards a particular type of scene. \\

\subsubsection{IBR}
\label{sec:traits_ibr}

\begin{figure}[!tb]
\centering
\includegraphics[width=8.5cm,keepaspectratio=true]{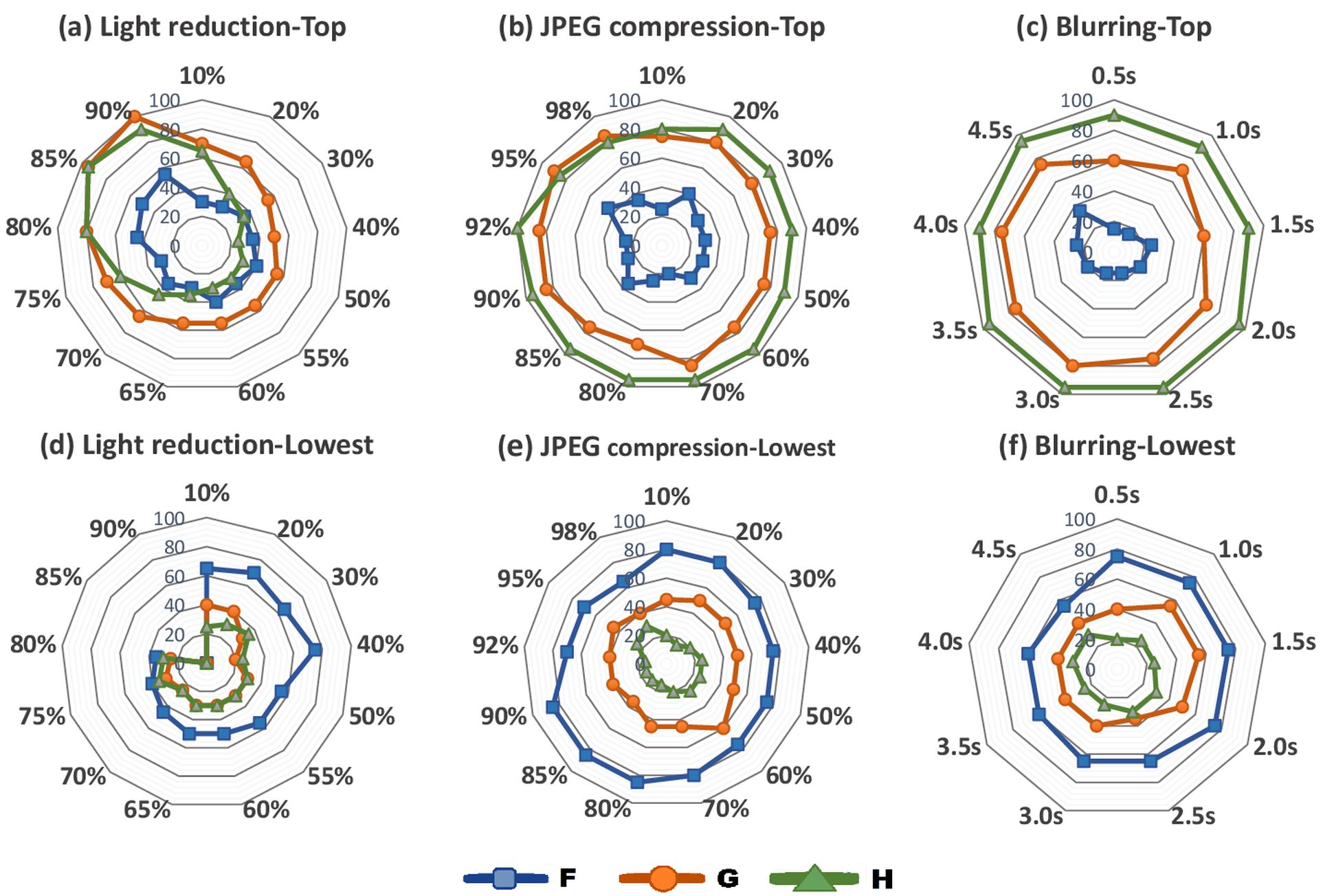}
\caption{
Top and lowest trait indices of IBR in percentage for different amounts of light reduction (a,d), JPEG compression (b,e) and blurring (c,f).
}
\label{fig:ibr_tables}
\end{figure}

\black{The uniform light change has a significant impact on the performance of IBR \cite{tuytelaars_matching_2004}\cite{ferrarini2016performance}. This made impossible to obtain the the lowest trait indices for brightness reduction at $85\%$ and $90\%$. Following the same approach as Section \ref{subsec:ebr_traits}, those indices are set to 0 (Figure \ref{fig:ibr_tables}.a).}
Under light reduction, the presence of a weak bias across the range of transformation amount is evident for human-made objects: $G$ indices are never below $50\%$ in top rankings while their counterpart in the lowest indices are never above $40\%$. A similar trend can be observed for $F$: the share of outdoor scenes in the top twenty is generally below $50\%$, while is generally never below $50\%$ for light reduction rates from $10\%$ to $65\%$.
Under JPEG compression, IBR achieved better performances on scenes, which are both simple and human made. Indeed, the related $G$ and $H$ indices in the top twenties reached very high values, which are never below $75\%$ and $80\%$ respectively (Figure \ref{fig:ibr_tables}.b).
The same kind of bias observed for JPEG compression characterized IBR under blurring as well: the top rankings are mainly populated by human made and simple scenes, whereas the lowest rankings contain mostly scenes with the opposite characteristics (Figure \ref{fig:ibr_tables}.c and \ref{fig:ibr_tables}.f). \\

\subsubsection{MSER}

\begin{figure}[!tb]
\centering
\includegraphics[width=8.5cm,keepaspectratio=true]{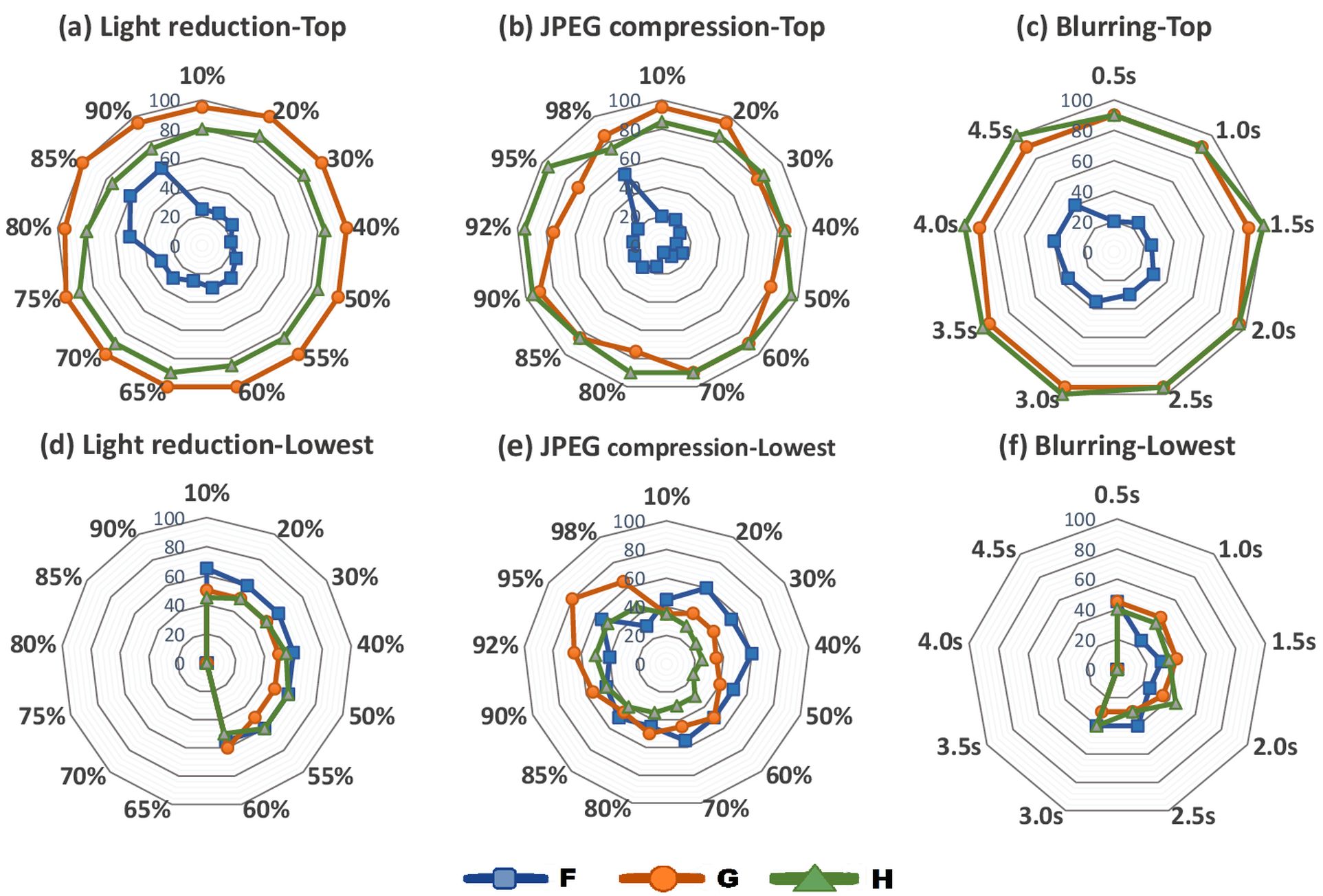}
\caption{
Top and lowest trait indices of MSER in percentage for different amounts of light reduction (a,d), JPEG compression (b,e) and blurring (c,f).
}
\label{fig:mser_tables}
\end{figure}

Figure \ref{fig:mser_tables} shows the trait indices for MSER \cite{matas_robust_2004}. Due to sensitivity of MSER to uniform light reduction and Gaussian blur, it has not been possible to compute the trait indices for the lowest rankings at light reductions of more than $60\%$ and for the last three steps of blurring as the number of scenes with repeatability equal to $0$ exceed the length of the lowest rankings at those transformation amounts.\\
\black{
The trait indices draw a very clear picture of the MSER's biases. The very high values of $G$ and $H$ of the top ranking indices and the relatively low values obtained for the lowest twenty rankings, lead to the conclusion that MSER performs significantly better on simple and human-made dominated scenes for every transformation type and amount.
}
Finally, the outdoor scenes populate mainly the lowest rankings built under light reduction and JPEG compression transformations while $F$ for blurring has low and balanced values between the top and lowest rankings. \\

\subsubsection{SALIENT}

\begin{figure}[!tb]
\centering
\includegraphics[width=8.5cm,keepaspectratio=true]{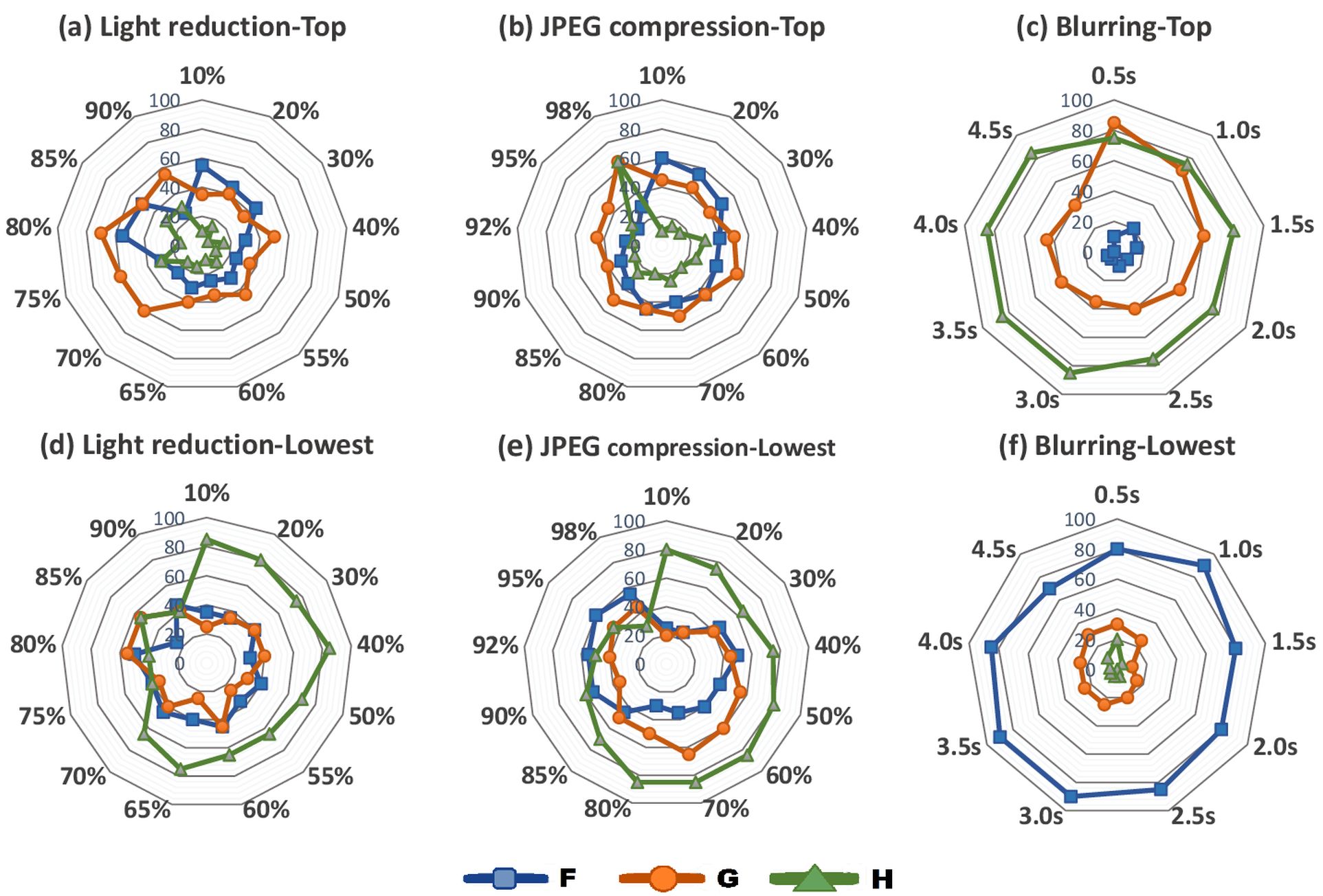}
\caption{
Top and lowest trait indices of SALIENT in percentage for different amounts of light reduction (a,d), JPEG compression (b,e) and blurring (c,f).
}
\label{fig:salient_tables}
\end{figure}

\black{The results for uniform light reduction (Figure \ref{fig:salient_tables}.a) show a strong preference of SALIENT \cite{kadir_affine_2004} for complex scenes as can be inferred by the low values of the index $H$ in the top twenties contrary to high values in the lowest rankings. This can be explained considering that uniform light reduction does not alter the shape of the edges and others lines present in a scene but makes low contrast region harder to  be identified.
Thus, the application of the uniform light transformation has the effect of populating the top ranking of SALIENT with those images containing a few but high contrast elements.}

On the other hand, the result for Gaussian blur (Figure \ref{fig:salient_tables}.c) shows a completely opposite situation, in which the most frequent scenes in the top rankings are those characterized by simple structure or, in other words, scenes whose information has relevant components at low frequencies. Indeed, as indicated above, Gaussian  blurring can be seen as a low pass filter and applying  it to an image results in loss of the information at high frequencies. Under JPEG compression, SALIENT exhibits a preference for complex scene as it is under light reduction with the difference that the $H$ indices increase with the compression rate. Although, JPEG compression is lossy and may alter the shape of the edges delimiting the potential salient region in an image, the impact on the information content is lower than the one caused by Gaussian blur. Indeed, the share of simple images is constantly low: $H$ below $30\%$ up to $98\%$. At $98\%$ the share of simple images in the top twenty increase dramatically to $65\%$ as the images lose a huge part of their information content due to the compression, which produce wide uniform regions (see the Figure \ref{fig:transformation_sample} for an example).\\

\subsubsection{SFOP}

\begin{figure}[!tb]
\centering
\includegraphics[width=8.5cm,keepaspectratio=true]{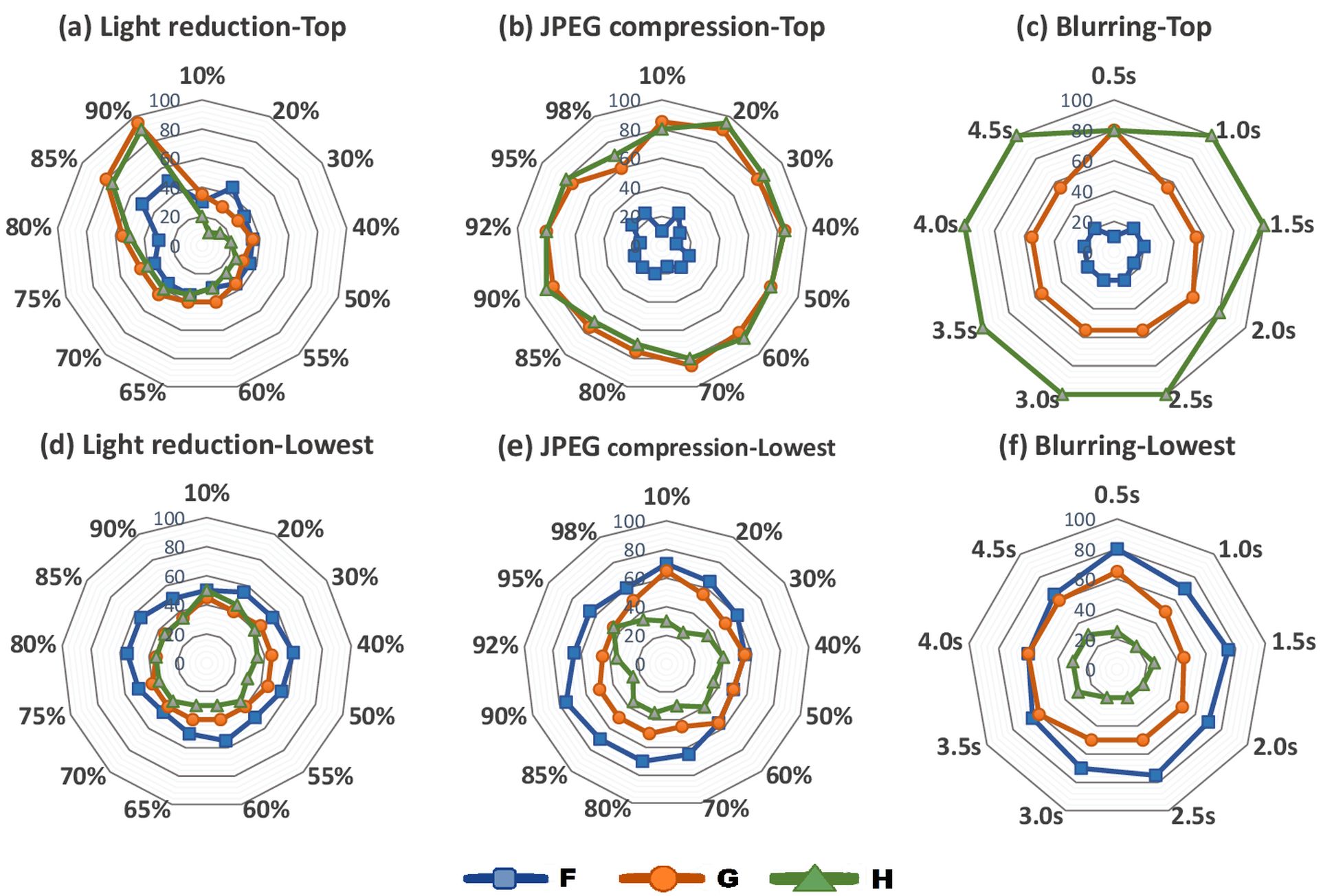}
\caption{
Top and lowest trait indices of SFOP in percentage for different amounts of light reduction (a,d), JPEG compression (b,e) and blurring (c,f).
}
\label{fig:sfop_tables}
\end{figure}

Under JPEG compression and Gaussian blur, the bias of SFOP \cite{forstner_detecting_2009} is towards simple scenes representing non-outdoor scenes. The kind of objects favored are human made under JPEG compression, while for blurring no clear preference can be inferred, due to closeness of the values of  $G$ indices between the top and lowest rankings. The relative measures of those biases are reflected by the $G$ and $H$ indices reported in the Figures \ref{fig:sfop_tables}.b and  \ref{fig:sfop_tables}.c: $H$ assumes high percentage values in the top rankings and low values in the lowest rankings; the indices $G$ for the top rankings of JPEG compression are constantly above $70\%$ whereas the related value registered for the lowest rankings exceeds $55\%$ only at $10\%$ of compression rate. The indices obtained for uniform light reduction reveal that SFOP performs lowest on outdoor scenes as can be seen by the lowest ranking $F$ values, which mostly fluctuate between $50\%$ and $60\%$. \\

\subsubsection{SURF}

\begin{figure}[!tb]
\centering
\includegraphics[width=8.5cm,keepaspectratio=true]{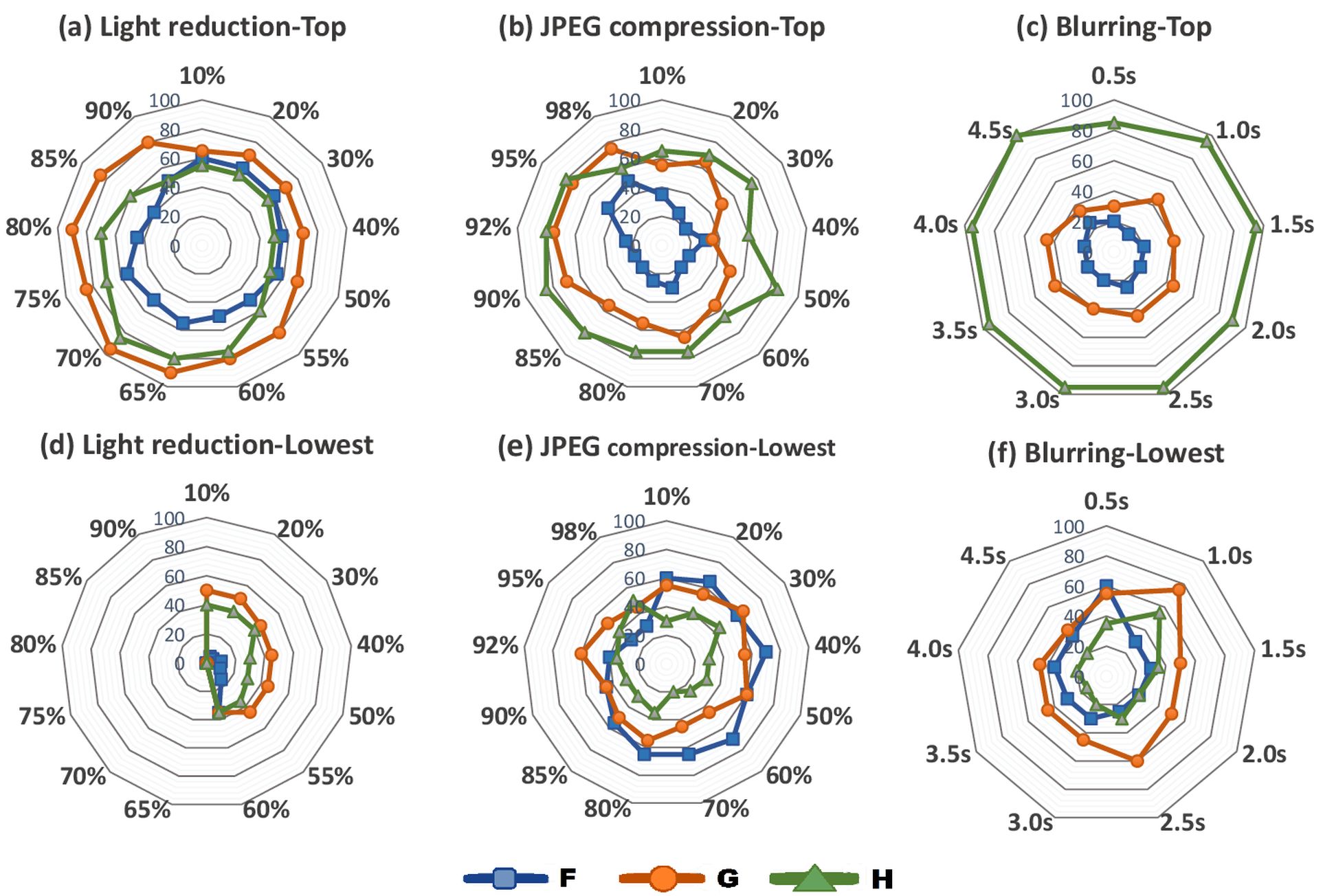}
\caption{
Top and lowest trait indices of SURF in percentage for different amounts of light reduction (a,d), JPEG compression (b,e) and blurring (c,f).
}
\label{fig:surf_tables}
\end{figure}

The performance of SURF is particularly affected by uniform light transformation and, because of that, it has not been possible to compute the trait indices at $65\%$ and further brightness reductions (Figure \ref{fig:surf_tables}.a). The effect of this transformation is to focus the biases of SURF towards human made objects ($G$ greater than or equal to $65\%$). Although the available $F$ indices for lowest rankings are extremely low (normally within $15\%$), only a weak bias towards outdoor scenes can be claimed as the highest values for $F$ indices in the top twenties are only $60\%$ between $10\%$ and $60\%$ of light reduction. The percentage of simple scenes in the top rankings fluctuate between $50\%$ and $85\%$ which, unfortunately, is reached in a region where the indices are not available thus, a comparison is not possible.\\
\black{
JPEG compression produces more predictable biases on SURF: $H$'s values are significantly higher in the top rankings that in the lowest rankings and the performance are worse with outdoor scenes than with non-outdoor scenes. Finally, the $G$ indices do not express a true bias, neither for human-made nor for natural elements.  (Figure \ref{fig:surf_tables}.d).}  \\
Under blurring SURF best performs on simple scenes whereas it performs poorer on complex scenes. The values of $F$ and $G$ for the top and lowest rankings are fairly close. $F$ values for both rankings groups are very low (except for $0.5\sigma$ which reaches $60\%$ for the lowest ranking) while $G$'s values fluctuate around $50\%$.

\section{Conclusions}
\label{sec:conclusions}

For several state-of-the-art feature detectors, the dependency of the repeatability from input scene type has been investigated utilizing a large database composed of images from a wide variety of human-classified scenes under three different types and amounts of image transformations. Although the utilized human-based classification method includes just three independently assigned labels, it is enough
to prove that the feature detectors tend to score their highest and lower repeatability scores with particular types of
scenes. \black{The detector preferences for a particular category of scene are pronounced and stable across the type and amount of image transformation for some detectors, such as MSER and EBR. Some detectors\rq{} bias are influenced more than others by the amount of transformation and the top-trait indices of SFOP under light changes are a good example: $G$ and $H$ reach a peak at $98\%$ of light reduction. In a few cases the indices show very similar values between top end lower rankings. This allows to conclude that biases of a detector are not sensitive to a particular image change.  For example, the trait indices of SALIENT for JPEG compression are between $40\%$ and $60\%$ for most of JPEG compression rate.\\
A significant number of local image feature detector have been assessed in this work, however the proposed framework is general and can be utilized for assessing any arbitrary set of detectors.} A  designer who needs to maximize the performance of a vision system starting from the choice of the better possible local feature detector could take advantage from the proposed framework. \black{Indeed, the framework could be utilized for identifying  the detectors which perform better with the type of scene most common in the application before any task-oriented evaluation (e.g. \cite{shin1998objective}, \cite{shin1999comparison}) thus, such a selection would be carried out on a smaller set of local feature detectors.} For example, for an application which deals mainly with indoor scenes, the detectors should be short-listed are HESAFF, HESLAP, HARAFF and HASAFF which have been proven to achieve their highest repeatability rate with non-outdoor scenes. On the other hand, if an application is intended for working in an outdoor environment, EBR should be one of the considered local feature detectors, especially under light reduction transformation.\\
In brief, the framework proposed permits to characterize the feature detector against the scene content and, at the same time, represent a useful tool for facilitating the design of those visual application which utilize a local feature detector stage.

\section*{Acknowledgment}

The authors would like to thank the anonymous reviewers for their helpful comments and suggestions. This work was supported by the EPSRC grant number EP/K004638/1 (RoBoSAS).

\ifCLASSOPTIONcaptionsoff
  \newpage
\fi

\end{document}